%% file: example.tex
\documentclass[11pt]{tea}

\usepackage[utf8]{inputenc}
\usepackage[T1]{fontenc}

\usepackage{amssymb}
\usepackage{hyperref}
\usepackage{xcolor}
\usepackage[most]{tcolorbox}
\usepackage{graphicx,verbatim}
\usepackage{booktabs}
\usepackage{multirow}
\usepackage{booktabs}
\usepackage{amsmath}
\usepackage{graphicx} 
\usepackage{pifont}   
\usepackage{xcolor}
\usepackage{makecell}
\usepackage[table]{xcolor}
\usepackage{tabularx}
\usepackage{fvextra}
\usepackage{array}
\usepackage{caption}
\usepackage{xspace}
\newcommand{\SlideBank}{\textbf{SlideBank}\xspace}
\tcbuselibrary{skins}

\definecolor{PromptGreen}{HTML}{1A531A}  
\definecolor{PromptGreenBg}{HTML}{F0F9F0}

\title{SlideBank: A Persistent Hierarchical Evidence Bank for \\Consistent Whole-Slide Reasoning}

\author[1,2,*]{Beidi Zhao}
\author[1,2,*]{Gexin Huang}
\author[3,*]{Ciro Zhang}
\author[4]{Anqi Li}
\author[5]{Yusheng Tan} 
\author[1,6]{Chen Zhou}
\author[1,6]{Gang~Wang} 
\author[1,6]{Zu-hua Gao} 
\author[1,2,\dagger]{Xiaoxiao Li} 

\affiliation[1]{University of British Columbia}
\affiliation[2]{Vector Institute}
\affiliation[3]{Harvard University}
\affiliation[4]{Rice University}
\affiliation[5]{University of Chicago}
\affiliation[6]{BC Cancer Agency}

\contribution{%
 $^*$ Equal contribution\\  $^\dagger$ Correspondence to Xiaoxiao Li at \href{mailto:xiaoxiao.li@ece.ubc.ca}{xiaoxiao.li@ece.ubc.ca}%
}

\abstract{
Whole-slide images (WSIs) are challenging for vision-language reasoning because diagnostically relevant morphology is sparse, heterogeneous, and distributed across gigapixel-scale images and multiple spatial resolutions. Existing WSI models and pathology agents can aggregate slide features or actively acquire evidence, but the information retained after exploration is often difficult to access semantically while preserving its connection to the original visual evidence. We introduce \SlideBank, a training-free framework that represents each WSI as a persistent, concept-indexed, and spatially grounded evidence bank. \SlideBank performs question-independent coarse-to-fine exploration to identify informative regions and multi-scale views, converts them into explicit morphological observations, and grounds pathology signals to their supporting patches and WSI coordinates. At inference time, questions are routed to relevant signals and evidence scales, and the linked global, regional, and patch evidence is integrated through confidence-based cross-level consensus. Experiments on WSI-VQA and SlideBench-BCNB show that 
with Patho-R1, SlideBank reaches 52.77\% on WSI-VQA and with Quilt-LLaVA, it reaches 50.92\% average accuracy on SlideBench-BCNB, while structured signal-guided retrieval consistently outperforms random evidence sampling. Reusing the same bank across repeated queries further achieves over 99\% rephrasing consistency and substantially reduces amortized inference cost through persistent evidence reuse.
}

\begin{document}
\maketitle
\input{sec/1_intro}

\input{sec/2_relatedwork}

\input{sec/3_method_new}

\input{sec/4_experiment}

\input{sec/5_conclusion}

\bibliographystyle{plainnat}
\bibliography{reference} 
\newpage
\appendix
\input{sec/appendix}

\end{document}

%% file: sec/1_intro.tex
\section{Introduction}
\label{sec:intro}
Whole-slide images (WSIs) are among the most challenging visual inputs in computational pathology.
Unlike conventional natural or medical images, a single WSI can contain billions of pixels and span tissue structures ranging from large-scale architectural patterns to fine-grained cellular morphology.
Diagnostically relevant findings are often sparse, spatially dispersed, and highly heterogeneous: a small focus of atypical cells, an invasive tissue interface, or a mitotic hotspot may occupy only a tiny fraction of the entire slide while being decisive for diagnosis.
Consequently, directly compressing a WSI into a thumbnail or uniformly processing a limited set of patches can easily discard critical evidence.
Effective WSI reasoning therefore requires not only understanding \emph{what} is visible, but also determining \emph{where} diagnostically informative evidence lies, \emph{at what scale} it should be examined, and \emph{how} evidence distributed across the slide should be organized for downstream reasoning.

Recent work has approached this challenge from two major directions.
WSI-level vision-language models~\cite{chen2025slidechat,wsillava2025iccv,ding2025multimodal,alawode2026mllm} aggregate precomputed patch or slide features into compact latent representations, enabling end-to-end reasoning over gigapixel images.
More recently, pathology agents~\cite{pathfinder2025arxiv,chen2025pathagent,yang2026pathnavigate} have treated WSI understanding as an active evidence-acquisition problem, using planning, region selection, and multi-scale navigation to locate diagnostically relevant fields before generating an answer.
Several methods further introduce hierarchical representations, cached observations, or slide-specific memory to improve efficiency and reasoning over large visual contexts~\cite{alawode2026mllm,chen2025pathagent,yang2026pathnavigate,huang2025survagent,li2026pathmem}.
These approaches have substantially advanced WSI understanding, but the information retained after slide exploration is typically represented either as latent visual features, flat collections of observations, or states optimized for a particular navigation or reasoning trajectory.
How to organize visually established evidence from the \emph{current slide} into an explicit representation that can be semantically accessed by different downstream questions remains comparatively underexplored.

This motivates the central question of this work:
\emph{how can evidence acquired from a gigapixel WSI be transformed into a persistent representation that supports diverse downstream questions while preserving explicit traceability to the underlying visual observations?}
This perspective shifts the focus from evidence acquisition alone to how
acquired evidence is organized and accessed after exploration.
A useful evidence representation should expose clinically meaningful findings in a form that can bridge different question formulations, preserve the multi-scale structure of pathology evidence, and maintain a traceable connection between semantic findings and the fine-grained areas and WSI locations that support them.

Constructing such a representation introduces several challenges.
First, exhaustive inspection is infeasible at gigapixel scale, yet aggressive sampling risks missing sparse or heterogeneous abnormalities; the system must therefore identify a compact but diagnostically diverse set of fields across both architectural and cellular scales.
Second, the language used in downstream questions does not necessarily match the language of raw visual observations.
Questions about nuclear grade, pleomorphism, or cellular atypia, for example, may rely on overlapping morphological signals even when their surface forms differ, making direct retrieval over free-form descriptions unreliable.
Third, clinically meaningful evidence is inherently multi-scale and spatially grounded: a semantic finding should remain linked not only to a textual description, but also to the anchor region, magnification, visual field, and WSI coordinates from which it was derived.
Finally, different questions may require different combinations of global context, regional architecture, and cellular detail, requiring an evidence interface that remains persistent while supporting query-adaptive composition.

To address these challenges, we propose \textbf{SlideBank}, an agent-driven framework that requires no task-specific training and converts each WSI into a \textbf{concept-indexed, spatially grounded hierarchical evidence bank}. To construct the bank, SlideBank performs question-independent coarse-to-fine exploration, using a global survey and category-aware sampling to identify diagnostically informative anchor regions and complementary architecture- and cell-level views. A pathology generative model converts the selected views into explicit morphological observations, which are subsequently normalized into clinically meaningful signals by a concept-guided pathology signal layer. Each signal is associated with its status, confidence, supporting anchors, multi-scale views, coordinates, and magnifications, establishing a traceable path from \emph{pathology signal} to \emph{supporting evidence}, \emph{visual field}, and ultimately \emph{WSI location}. At inference time, a signal-aware reader routes each question to a relevant pathology concept, uses the associated signals to retrieve supporting anchors and multi-scale views, and generates candidates independently at the global, anchor, and patch levels. These candidates are combined through confidence-based selection and cross-level consensus. In this way, SlideBank explores each slide once, organizes the observed morphology into a persistent evidence bank, and reuses the same bank to answer different downstream questions.

Our contributions are three-fold:
\begin{itemize}
    \item We introduce \textbf{SlideBank}, a structured, slide-specific hierarchical evidence representation that organizes observed morphology through pathology signals while preserving explicit links to supporting regions, multi-scale views, and WSI coordinates.

    \item We develop a concept-to-signal retrieval method that couples concept-level question routing with signal-grounded evidence retrieval, providing a consistent semantic interface for indexing and reasoning across diverse question formulations while retaining traceability to the underlying visual evidence.

    \item We evaluate \textbf{SlideBank} on two public WSI question-answering benchmarks in terms of accuracy, efficiency, and cross-turn consistency. A controlled comparison using the same evidence bank shows the benefit of concept-to-signal retrieval over random evidence sampling.
\end{itemize}

\begin{figure*}[t]
    \centering
    \includegraphics[width=\linewidth]{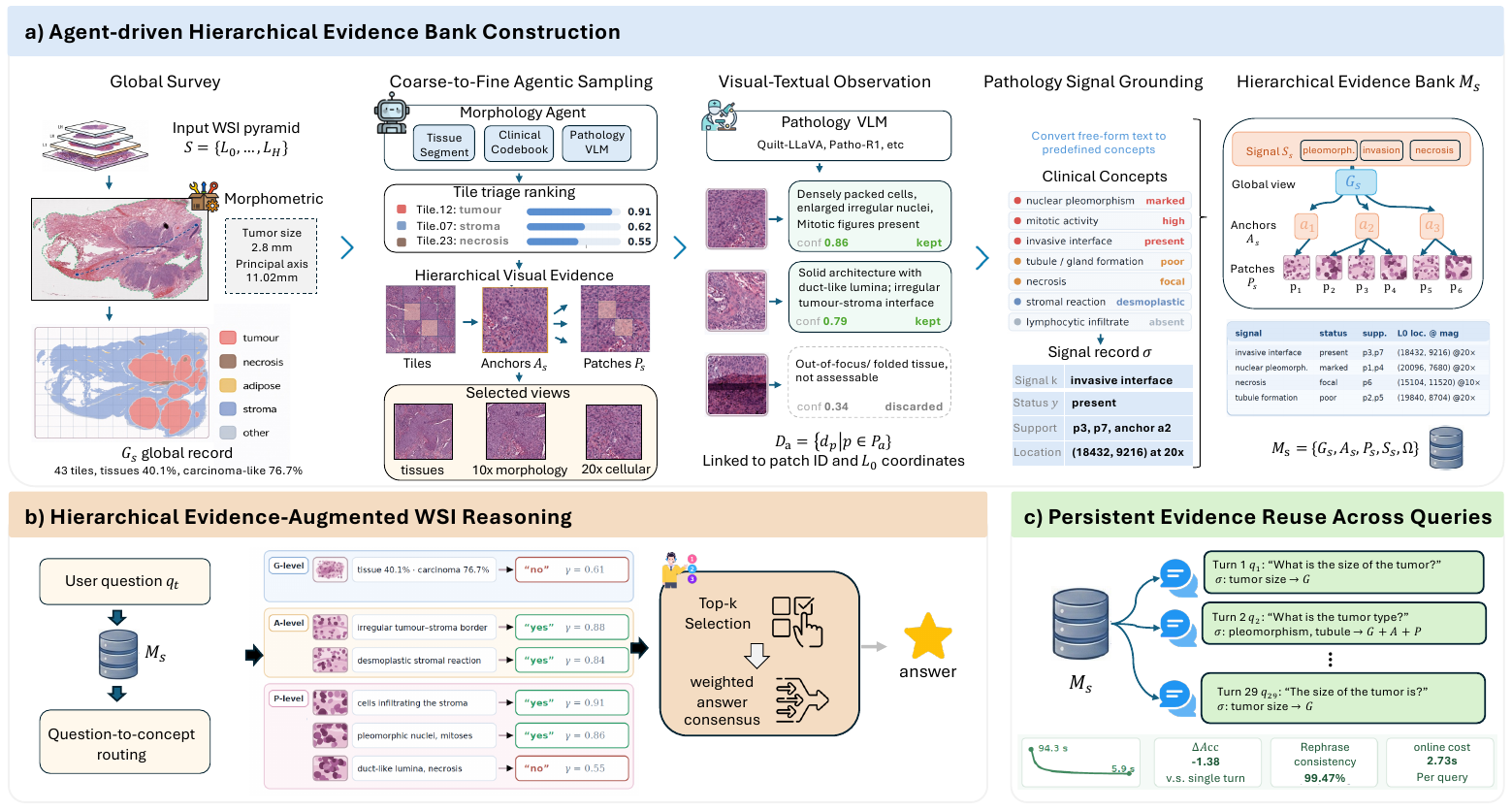}
    \caption{Overview of \textbf{SlideBank}.
(a) A pathology agent explores the WSI and organizes multi-scale observations into a spatially grounded evidence bank.
(b) At inference time, each question is routed to relevant pathology concepts and linked global-, anchor-, and patch-level evidence, whose predictions are combined through confidence-based selection and weighted cross-level consensus. 
(c) The same bank is reused across rephrased queries for stable and efficient reasoning.}
    \label{fig:framework}
\end{figure*}

%% file: sec/2_relatedwork.tex
\section{Related Work}
\label{sec:related}
\subsection{Vision--Language Models and Agentic Reasoning for Whole-Slide Pathology}
Early pathology vision--language models mainly learned image--text alignment or instruction following from localized fields. CONCH learns transferable visual--language representations, while Quilt-LLaVA and PathChat support instruction following and open-ended interaction with histology images~\cite{lu2024visual,seyfioglu2024quilt,lu2024pathchat}. Because these models cannot directly process a complete WSI at native resolution, subsequent methods aggregate pre-extracted patch features into slide-level representations. WSI-VQA introduced generative question answering over WSIs, TITAN learned a multimodal whole-slide representation, and SlideChat and WSI-LLaVA connected slide features to language models for slide-level dialogue and reasoning~\cite{chen2024wsi,ding2025multimodal,chen2025slidechat,wsillava2025iccv}. More recent systems improve hierarchical or large-scale slide understanding: MLLM-HWSI aligns cell-, patch-, region-, and slide-level tokens with pathology language, while ALPaCA and PRISM2 scale slide-level supervision through question answering and clinical dialogue~\cite{alawode2026mllm,alpaca2026naturecomm,prism22026naturemed}. In parallel, agentic systems make evidence acquisition more adaptive. PathFinder, WSI-Agents, SlideSeek, GIANT, and PathAgent introduce specialized agents, planning, tool use, iterative navigation, or evidence synthesis for gigapixel slides~\cite{pathfinder2025arxiv,lyu2025wsi,slideseek2025arxiv,giant2025arxiv,chen2025pathagent}; HistoSelect performs question-guided coarse-to-fine selection, BEACON acquires patches according to expected information gain, and AdaptivePath learns question-agnostic multiscale navigation~\cite{histoselect2026cvpr,beacon2026arxiv,adaptivepath2026arxiv}. EviPathBench further shows that locating diagnostic regions remains substantially harder than reasoning over preselected evidence~\cite{evipathbench2026arxiv}. PathNavigate is particularly related to our setting because it performs a question-independent scan and maintains an online memory over frozen slide features before question-conditioned search~\cite{yang2026pathnavigate}. However, existing methods primarily represent acquired evidence as model-internal features or transient reasoning states. \SlideBank instead materializes selected multiscale image fields, morphological descriptions, pathology signals, and WSI coordinates as an independently inspectable bank that can be retrieved across multiple questions about the same slide.

\subsection{Hierarchical Evidence and Memory}
Hierarchical modeling provides a natural way to represent gigapixel WSIs. HIPT learns nested patch- and region-level representations, while H$^2$-MIL organizes heterogeneous instances into a hierarchy for slide analysis~\cite{chen2022scaling,hou2022h}; however, both primarily structure latent visual features for downstream prediction. Concept bottleneck models expose human-interpretable intermediate variables, and post-hoc variants add such structure to pretrained models~\cite{koh2020conceptbottleneck,yuksekgonul2023posthoc}. Persistent memory has also been studied in long-horizon agents: Generative Agents store episodic experiences, MemGPT manages context through external memory, and A-MEM and G-Memory organize memories through linked or hierarchical structures~\cite{park2023generative,packer2023memgpt,xu2025mem,zhang2025g}. Related ideas have recently appeared in pathology. SurvAgent constructs a multimodal case bank for cross-case survival prediction, whereas PathMem organizes structured diagnostic knowledge as long-term memory and transfers relevant content into working memory~\cite{huang2025survagent,li2026pathmem}. These approaches focus on latent hierarchies, general episodic memory, cross-case experience, or domain knowledge rather than persistent spatial evidence from repeated examination of one WSI. \SlideBank combines hierarchical visual evidence with an interpretable routing interface: concepts represent coarse question-side categories, signals represent localized image-derived findings, and a predefined ontology maps concepts to relevant signals. Each signal record retains its status and links to associated anchors, multiscale patch views, descriptions, and WSI coordinates.The resulting slide-specific bank is reused across questions, so every turn is answered from the persistent evidence bank, not only affected by the accumulated conversation history.

%% file: sec/3_method_new.tex
\section{Method}
\label{sec:method}

\subsection{Overview}

Whole-slide reasoning is challenging since a gigapixel WSI exceeds the accepted maximum image size of current vision-language models (VLMs) and clinically relevant findings may occupy only a small fraction of the slide. \SlideBank therefore separates slide exploration from question answering. As illustrated in Fig.~\ref{fig:framework}, the framework consists of three components: \textbf{Agent-Driven Hierarchical Evidence Bank Construction} organizes multi-scale slide observations into a persistent, spatially grounded evidence bank; \textbf{Hierarchical Evidence-Augmented WSI Reasoning} retrieves and integrates question-relevant evidence from the bank; and \textbf{Persistent Evidence Reuse Across Queries} enables the same slide representation to support subsequent questions without reconstructing the underlying evidence.

Let a whole slide image be represented by a resolution pyramid $s=\{L_0,\ldots,L_H\}$, where $L_0$ and $L_H$ denote the highest- and lowest-resolution views, respectively. In Section~\ref{sec:bank}, \SlideBank performs question-independent coarse-to-fine exploration to construct a slide-specific evidence bank $M_s$ that organizes multi-scale observations and links pathology signals to their supporting evidence and WSI locations. In Section~\ref{sec:reasoning}, the online reader maps a question $q_t$ to relevant signals and evidence scales, retrieves the linked evidence, and combines global, regional, and patch-level predictions through confidence-based consensus. Finally, Section~\ref{sec:multiturn} reuses the same $M_s$ across subsequent queries.

\subsection{Agent-Driven Hierarchical Evidence Bank Construction}
\label{sec:bank}

\noindent\textbf{Global Survey and Coordinate Normalization.} 
To capture slide-level contextual cues and facilitate global morphological measurements, such as tumor size and disease extent, we first construct a global \textit{thumbnail}. For pyramidal WSIs, we use the lowest-resolution level; for non-pyramidal or single-resolution slides, we construct an equivalent bounded thumbnail via chunk-wise subsampling. All spatial coordinates identified on the thumbnail are proportionally mapped back to the level-0 coordinate system of the source WSI.
Foreground tissue is separated from the glass background using HSV thresholding, and the valid tissue area is partitioned into a non-overlapping $384\times384$ grid. Each tile is characterized using handcrafted morphological statistics and foreground occupancy, and assigned to one of five coarse tissue classes: tumor, necrosis, adipose, stroma, or other. When physical pixel spacing is available, we additionally estimate the maximum Feret diameter and principal axis of the primary tumor candidate. These slide-level observations, including the natural-language summary, tissue-composition statistics, and geometric diagnostics, are stored in a global record $G_s$ for slide $s$.

\noindent\textbf{Coarse-to-f{}ine Agentic Sampling.}
Tiles are first ranked according to coarse diagnostic priority, triage confidence, and tissue coverage, and retained under a category-aware budget that allocates more samples to suspicious regions while preserving coverage of the remaining tissue classes. Sampling then proceeds through recursive $4\times4$ refinement. For each selected tile, the agent scores all 16 cells of its downsampled representation to propose \textit{anchor regions}. Within each anchor region, the same procedure is repeated to identify informative $20\times$ and $10\times$ \textit{patches}. At both stages, scoring is guided by a fixed set of pathology-derived cues provided in the prompt (prompts used throughout \SlideBank are provided in Appendix).

\noindent\textbf{Multi-Scale Visual--Textual Observations.}
The resulting hierarchy separates coarse triage from fine-grained evidence: tiles provide the coarse sampling prior, whereas \textit{anchor regions} and their associated patches provide the observed morphological evidence. For each \textit{patch} $p$, a pathology-specific VLM generates a concise morphological description $d_p$, together with a self-reported confidence score and a quality-control flag, conditioned on the \textit{patch} image and its magnification. Descriptions with confidence below a predefined threshold or invalid quality-control flags are discarded. 

For each retained anchor region $a$, the descriptions of its multi-magnification patches are collected as
\begin{equation}
    \mathcal{D}_a = \{\, d_p \mid p \in \mathcal{P}_a \,\},
    \label{eq:anchor-desc}
\end{equation}
where $\mathcal{P}_a$ denotes the patches associated with anchor $a$.
This anchor-level textual observation summarizes complementary morphology across magnifications while retaining links to the originating patches and their WSI coordinates.
We write $\mathcal{A}_s$ for the set of \textit{anchor regions} retained for slide $s$, and
$\mathcal{P}_s = \bigcup_{a \in \mathcal{A}_s} \mathcal{P}_a$
for all multi-scale \textit{patch} views collected under them; each $p \in \mathcal{P}_s$ stores its image, magnification, level-0 coordinates, and description $d_p$.

\noindent\textbf{Pathology Signal Grounding.}
Free-form descriptions may express the same finding in different ways. We therefore map $\mathcal{D}_a$ to a predefined vocabulary of localized pathology \emph{signals}, such as marked nuclear pleomorphism, high mitotic activity, low tubule formation, and tumor necrosis. For each applicable signal $k$, we store
\begin{equation}
\sigma_{a,k}
=
\left(
y_{a,k},
\rho_{a,k},
\mathcal{P}^{+}_{a,k}
\right),
\end{equation}
where $y_{a,k}\in\{\mathrm{present},\mathrm{absent},\mathrm{unknown}\}$ is the inferred status, $\rho_{a,k}$ is its confidence, and $\mathcal{P}^{+}_{a,k}\subseteq\mathcal{P}_a$ contains the \textit{patch} views supporting a positive assessment. This standardized signal layer preserves links to the underlying descriptions, images, and WSI locations.

\noindent\textbf{Persistent Hierarchical Storage.}
The signal records across all anchors are collected as
\begin{equation}
    \mathcal{S}_s = \{\, \sigma_{a,k} \mid a \in \mathcal{A}_s,\ k \in \mathcal{K}_a \,\},
    \label{eq:signal-set}
\end{equation}
where $\mathcal{K}_a$ is the set of signals applicable to anchor $a$.
We also define an ontology $\Omega$ that maps each coarse pathology concept $c \in \mathcal{C}$ to a set of relevant signals $\mathcal{K}(c)$.
The slide bank is then represented as
\begin{equation}
    M_s = \bigl(G_s,\ \mathcal{A}_s,\ \mathcal{P}_s,\ \mathcal{S}_s,\ \Omega\bigr),
    \label{eq:bank}
\end{equation}
comprising the global record, the retained anchor regions, their multi-scale \textit{patch} views, the grounded signal records, and the concept-to-signal ontology. The complete pathology concept vocabulary, concept-to-signal mapping and evidence-bank schema are provided in Appendix.

\subsection{Hierarchical Evidence-Augmented WSI Reasoning}
\label{sec:reasoning}
\noindent\textbf{Question-to-Concept Routing and Signal Retrieval.}
A pathology \emph{concept} represents the coarse intent of a question, such as tumor grading, necrosis assessment, or margin status, whereas a pathology \emph{signal} represents a localized image-derived finding.
Given a question $q$ and its options $O$, a deterministic router $\psi$ returns the top-$n$ pathology concepts
\begin{equation}
    \mathcal{C}_q = \psi(q, O), \quad |\mathcal{C}_q| = n,
    \qquad
    \mathcal{K}_q = \bigcup_{c \in \mathcal{C}_q} \mathcal{K}(c).
    \label{eq:routing}
\end{equation}
$\psi$ is based on lexical rules over the pathology concept vocabulary $\mathcal{C}$ (details in Appendix). We use $n=3$ by default, sensitivity to n is analyzed in Sec.~\ref{sec:sensitivity}.
The target signals are then used to retrieve
\begin{equation}
    \mathcal{S}(q) = \{\, \sigma_{a,k} \in \mathcal{S}_s \mid k \in \mathcal{K}_q \,\}.
    \label{eq:signal-retrieval}
\end{equation}
For each target signal, \SlideBank selects the \textit{anchor regions} in which that signal is most strongly supported.
Records labeled as \texttt{present} are preferred over uncertain or absent records, and records with the same status are ranked by confidence.
The associated anchor images, \textit{patch} views, and morphological descriptions are then retrieved as the question-specific evidence $E(q) = (G_s, \mathcal{A}_q, \mathcal{P}_q, \mathcal{S}_q)$.

\noindent\textbf{Evidence-Conditioned Branch Predictions.}
The question-specific evidence
$E(q)=(G_s,\mathcal{A}_q,\mathcal{P}_q,\mathcal{S}_q)$
is organized into three complementary branches
$\ell\in\{G,A,P\}$. The global branch captures slide-wide context, the
anchor branch represents regional architecture, and the \textit{patch} branch
provides localized morphology. Anchor and \textit{patch} inputs include their stored
morphological descriptions, whereas the global branch uses the slide
thumbnail without a local description.

Each evidence item is evaluated independently by the inference VLM. Let
$\mathcal{O}$ denote the answer-option set and
$z_{\ell,i}(o)$ the logit assigned to option $o\in\mathcal{O}$ for item $i$
from branch $\ell$. Its option probability is
\begin{equation}
\pi_{\ell,i}(o)
=
\frac{\exp z_{\ell,i}(o)}
{\sum_{o'\in\mathcal{O}}\exp z_{\ell,i}(o')}.
\end{equation}
The corresponding prediction and confidence are
\begin{equation}
\hat{o}_{\ell,i}
=
\arg\max_{o\in\mathcal{O}}\pi_{\ell,i}(o),
\qquad
\gamma_{\ell,i}
=
\max_{o\in\mathcal{O}}\pi_{\ell,i}(o).
\end{equation}
Within the anchor and \textit{patch} branches, candidates are ranked by
$\gamma_{\ell,i}$, and the top-$m$ candidates are retained as
$\mathcal{T}_{\ell}(q)$.

\noindent\textbf{Level-Weighted Answer CSonsensus.}
For each local branch $\ell\in\{A,P\}$, the retained candidates are
aggregated by majority voting. The resulting support for option $o$ is
\begin{equation}
v_{\ell}(o\mid q)
=
\frac{1}{|\mathcal{T}_{\ell}(q)|}
\sum_{i\in\mathcal{T}_{\ell}(q)}
\mathbb{I}\!\left[\hat{o}_{\ell,i}=o\right].
\end{equation}
For the global branch, the option probabilities are used directly:
\begin{equation}
v_G(o\mid q)=\pi_{G,1}(o).
\end{equation}
The three branches are combined using
\begin{equation}
\label{eq:fusion}
F(o\mid q)
=
\sum_{\ell\in\{G,A,P\}}
\lambda_{\ell}v_{\ell}(o\mid q),
\qquad
\sum_{\ell}\lambda_{\ell}=1,
\end{equation}
and the final answer is
\begin{equation}
\hat{o}(q)
=
\arg\max_{o\in\mathcal{O}}F(o\mid q).
\end{equation}
We use $\lambda_G=0.2$, $\lambda_A=0.4$, and $\lambda_P=0.4$,
placing greater emphasis on localized morphological evidence while retaining
regional and global context. The retained candidates, signal records, and
spatial evidence links provide a trace from the final answer to its
supporting WSI regions.

\subsection{Persistent Evidence Reuse Across Queries}
\label{sec:multiturn}
\SlideBank constructs the evidence bank $M_s$ once and reuses it across all turns. For each question $q_t$, the reader performs question-dependent routing to obtain evidence $E(q_t)$ from the bank and predicts the answer with the evidence, avoiding repeated WSI exploration and reducing evidence drift. Routing depends only on the current question, so the retrieved evidence for a given question is identical whether it is asked in isolation or within a conversation.

To preserve conversational continuity without accumulating a long history, the reader carries only the previous question--answer pair,
\begin{equation}
    H_t = \bigl(q_{t-1}, \hat{o}(q_{t-1})\bigr),
\end{equation}
which is supplied to the inference VLM alongside the retrieved evidence. The visual evidence therefore always comes from the persistent bank, while $H_t$ provides bounded dialogue context.

%% file: sec/4_experiment.tex
\section{Experiments}

\noindent\textbf{Datasets.}
We evaluate \SlideBank on two public WSI-level benchmarks:
\textbf{WSI-VQA}~\cite{chen2024wsi}, which contains 85 WSIs with
slide-level visual question answering annotations, and
\textbf{SlideBench-BCNB}~\cite{chen2025slidechat}, which contains
1,058 breast cancer slides covering tumor typing, grading, and molecular
subtyping. The two benchmarks provide complementary settings for evaluating
general WSI question answering and clinically relevant diagnostic reasoning.
For repeated-query evaluation, we additionally construct a multi-turn setting
from WSI-VQA by grouping all questions associated with the same slide into one
conversation, resulting in an average of 29 questions per WSI. We further
generate five semantically equivalent formulations of each question with
shuffled answer options using Gemini-2.5-Flash~\cite{comanici2025gemini} to evaluate the
stability of predictions under different question formulations.

\noindent\textbf{Baselines.}
We compare against three families of methods.
\emph{Zero-shot VLMs} include general-purpose and pathology-specific models
that receive only a $1024\times1024$ WSI thumbnail.
\emph{WSI-trained models} include WSI-LLaVA, TITAN, and SlideChat, which
are trained or adapted for slide-level understanding.
\emph{Agentic systems} include Med-Agents and PathAgent, which
introduce additional inference-time reasoning or WSI exploration.
All methods are evaluated on the same question sets and answer spaces.
Thumbnail-based baselines receive no additional high-resolution patches.
We evaluate \SlideBank with Quilt-LLaVA and Patho-R1 to examine
backbone transferability.

\noindent\textbf{Metrics.}
For standard WSI QA, we report accuracy on WSI-VQA and question-count-weighted average over the three BCNB tasks (tumor, histological grading and molecular subtype). For repeated-query evaluation, we
additionally report accuracy, the change from single-query accuracy
($\Delta$Acc), and rephrase consistency ($
    \mathrm{Cons}_i =
    \mathbb{I}\left[
        \max_{a}\sum_{j=0}^{5}
        \mathbb{I}\bigl(\hat{a}_i^{(j)}=a\bigr)
        \geq 3
    \right],
$) across semantically equivalent
question formulations.

\noindent\textbf{Implementation Details.}
Within each experimental setting, we use the same pathology VLM backbone for both evidence-bank construction and online inference, without additional training. We show the result with Quilt-LLaVA \cite{seyfioglu2024quilt} and Patho-R1 \cite{zhang2025pathor1} as the backbone. For evidence-bank construction, we partition the slide thumbnail into $384\!\times\!384$ tiles, retain those with at least $20\%$ tissue coverage, and perform both tile-level anchor localization and within-anchor patch selection on a $4\!\times\!4$ grid. We select three anchors from each tumor-like tile and two from each retained non-tumor tile. For each anchor, we extract an adaptive context crop capped at 4096 level-0 pixels, a 512-pixel fine-detail view corresponding to approximately $20\times$ on slides scanned at $40\times$, and an additional architectural view at approximately $10\times$. All views are resized to $256\!\times\!256$ for VLM processing. During online inference, we use all retrieved anchor-level evidence and sample at most 32 patch-level evidence items with a fixed random seed 42. We retain the top five ($m$=5) candidates from the anchor and patch branches according to token-probability confidence and fuse the global, anchor, and patch predictions using $(\lambda_G,\lambda_A,\lambda_P)=(0.2,0.4,0.4)$. All experiments use a single NVIDIA A100 GPU with 80\,GB memory, with FP16 inference for Quilt-LLaVA and BF16 inference for Patho-R1. Additional dataset, baseline, multi-turn evaluation, and answer-parsing details are provided in Appendix.
\input{tables/baseline}

\subsection{Standard WSI Question Answering}
Table~\ref{tab:baseline} evaluates whether the question-independent
evidence bank preserves sufficient slide information for downstream WSI
reasoning. Despite requiring no task-specific training, \SlideBank achieves
strong performance on both benchmarks. On SlideBench-BCNB, \SlideBank with
Quilt-LLaVA obtains 89.82\% accuracy on tumor typing and the highest overall
average of 50.92\%, substantially improving over the thumbnail-only
Quilt-LLaVA backbone. On WSI-VQA, \SlideBank with Patho-R1 reaches 52.77\%,
outperforming its thumbnail-only counterpart by 8.49 percentage points and
achieving the best performance among the evaluated methods. These results
show that organizing gigapixel WSI observations into a reusable structured
evidence bank preserves and can improve standard single-query reasoning performance.

\subsection{Effect of Evidence Organization and Access}
\noindent\textbf{Multi-scale Hierarchy.}
We first examine how different levels of the evidence hierarchy contribute
to WSI reasoning. As shown in Table~\ref{tab:ablation}, local evidence is
substantially more informative than the global thumbnail alone. On WSI-VQA,
the anchor and patch branches achieve 51.98\% and 51.45\% accuracy,
respectively, compared with 48.55\% using only global evidence. A similar
trend is observed on SlideBench-BCNB, where anchor-level evidence provides
the strongest single-level performance at 50.23\%. Combining global, anchor,
and patch evidence shows the best overall results on both benchmarks,
reaching 52.77\% on WSI-VQA and 50.36\% on SlideBench-BCNB.

\input{tables/ablation}

\noindent\textbf{Signal-guided Access.}
We next show whether the gain arises from structured evidence access rather
than simply exposing the model to more evidence. Using the same
constructed evidence bank, we replace signal-guided retrieval with uniformly sampling 32 patches at random while keeping other settings unchanged. Signal-guided retrieval improves
accuracy from 50.13\% to 52.77\% on WSI-VQA and from 49.62\% to 50.36\% on
SlideBench-BCNB. The
improvement shows the benefit of organizing and accessing slide evidence
through pathology signals compared to random sampling.

\subsection{Sensitivity and Qualitative Example}
\label{sec:sensitivity}
\noindent\textbf{Sensitivity of the Number of Concepts.}
\begin{figure}[t]
    \centering
    \includegraphics[width=0.5\columnwidth]{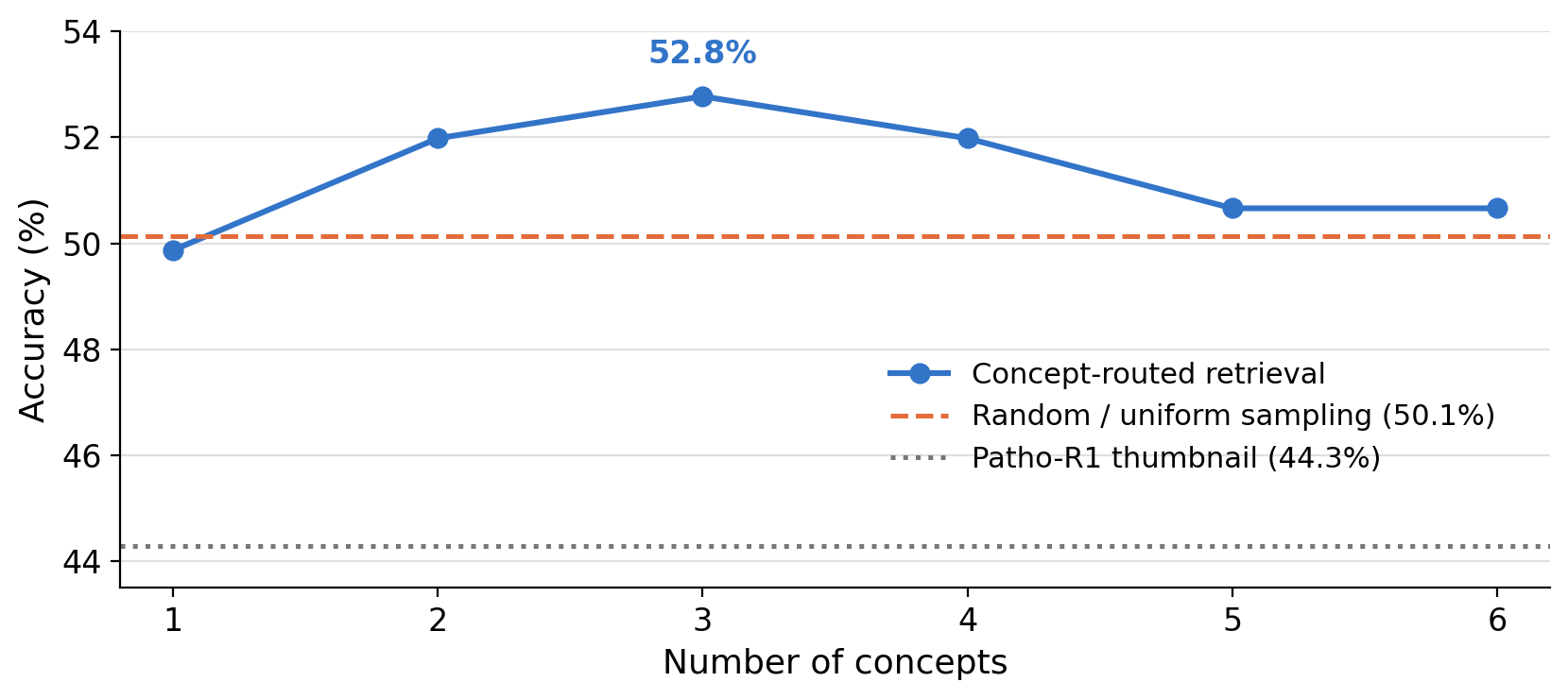}
    \caption{Sensitivity to the number of pathology concepts used for concept-routed retrieval (w/ Patho-R1).}
    \label{fig:concept}
\end{figure}
We study the sensitivity to the number of pathology concepts used for
concept-routed retrieval. As shown in Fig.~\ref{fig:concept}, accuracy
increases from 49.9\% with a single concept to a peak of 52.8\% with three
concepts, and then gradually decreases as additional concepts are included.
This reflects a trade-off between evidence coverage and retrieval noise:
too few concepts may omit relevant pathology signals, whereas overly broad
routing introduces less informative evidence. Performance remains above
random sampling across a broad range of 2--6 concepts, indicating that the
method is not sensitive to a narrowly tuned setting. We therefore use three
concepts by default.

\begin{figure*}[t]
    \centering
    \includegraphics[width=\linewidth]{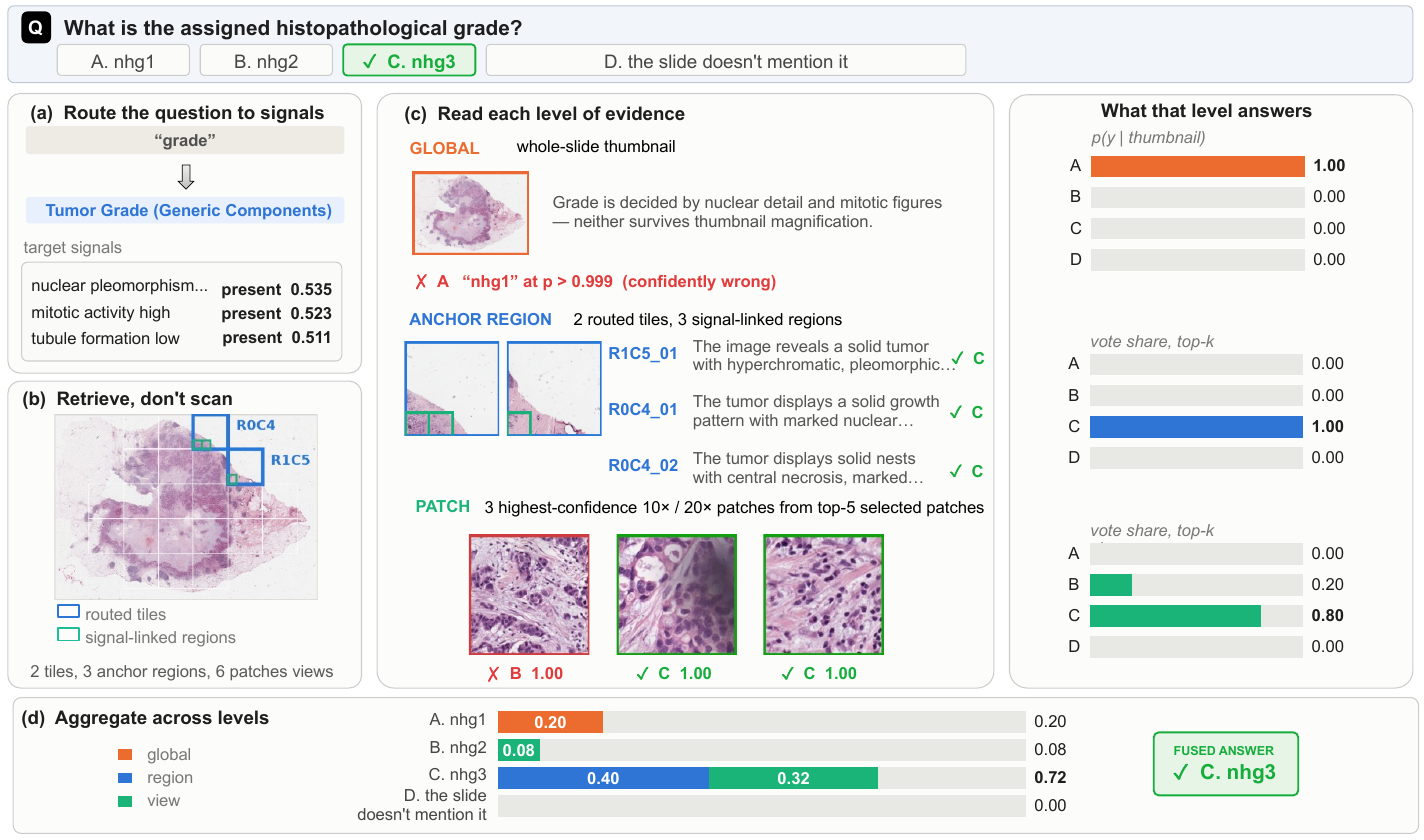}
    \caption{\textbf{Qualitative evidence trace in \SlideBank.}
Routed pathology signals retrieve supporting multi-scale evidence, which is fused to produce the final prediction.}
\vspace{-0.5cm}
    \label{fig:qualitative}
\end{figure*}
\noindent\textbf{Qualitative Example.}
Figure~\ref{fig:qualitative} illustrates the complete evidence trace for a
representative grading question. The query is first routed to grade-related
signals, which retrieve their linked regions and multi-magnification
patches. This example demonstrates how \SlideBank preserves an explicit
path from the final prediction back to the pathology signals and their
supporting visual evidence.

\begin{figure}[t]
    \centering
    \includegraphics[width=0.5\columnwidth]{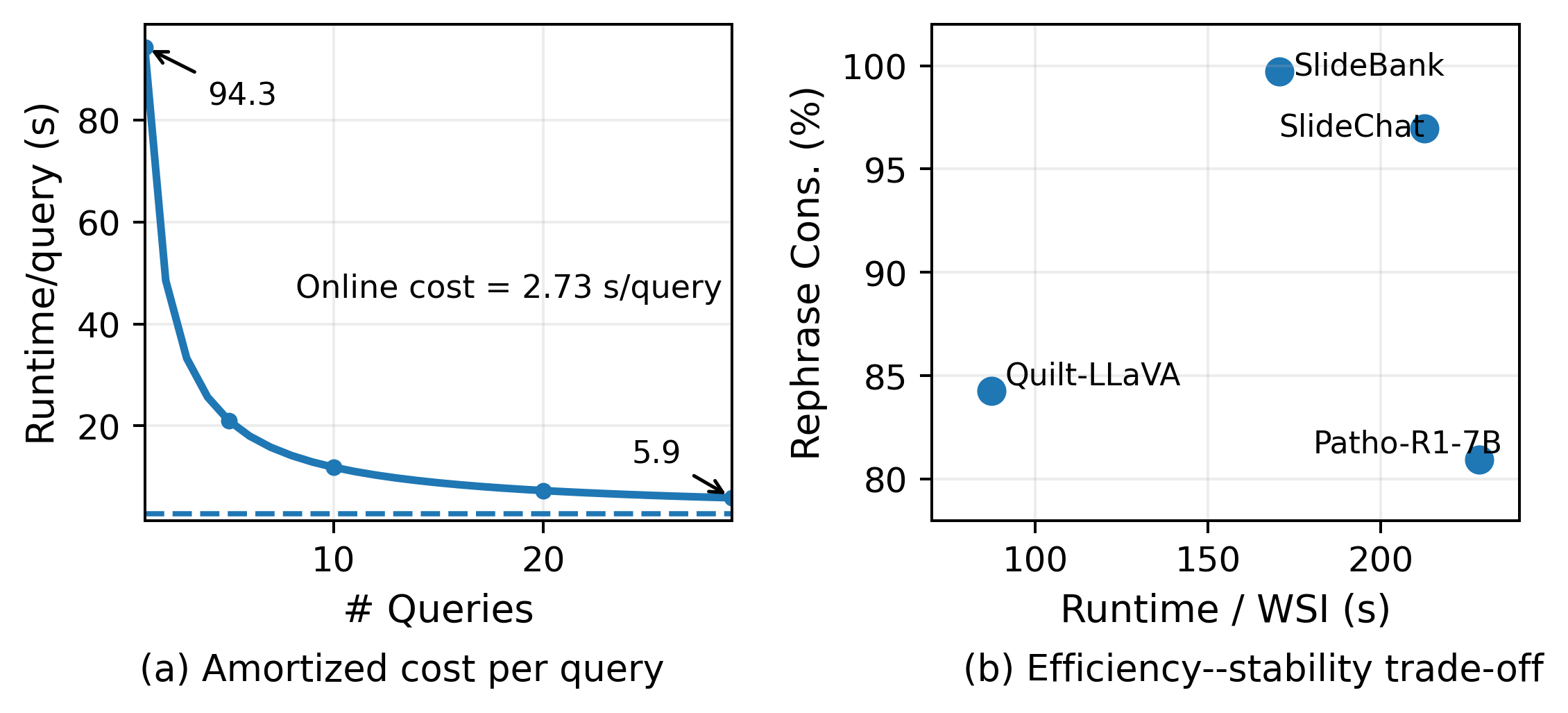}
    \caption{Efficiency and stability under repeated WSI queries.
(a) Amortized runtime per query when reusing the evidence bank (w/ Quilt-LLaVA).
(b) Runtime and consistency comparison among high-consistency methods.}
    \label{fig:efficiency}
\end{figure}

\subsection{Persistent Evidence Reuse Across Queries}
\noindent\textbf{Rephrased-QSuery Stability.}
We evaluate whether a persistent slide-specific evidence bank provides a stable basis for answering repeated questions about the same WSI. We group all WSI-VQA questions associated with each slide into a single sequence, leading to 29 questions per WSI on average, and present each question in a \textit{rephrased form with shuffled options}. As shown in Table~\ref{tab:multiturn}, several baselines suffer substantial accuracy degradation compared with independent single-query inference. In contrast, \SlideBank shows only modest drops of 1.38 and 3.83 percentage points with Quilt-LLaVA and Patho-R1, respectively, while achieving 99.47\% and 99.21\% consistency across semantically equivalent question formulations. Since routing depends only on the current question, these small drops arise from question rephrasing and the bounded dialogue context rather than from re-exploring the slide like baselines.
\input{tables/consistency}

\noindent\textbf{Amortized Efficiency.}
Figure~\ref{fig:efficiency} evaluates the computational benefit of reusing
the same evidence bank across repeated queries. Although bank construction
introduces a one-time cost, this cost is progressively amortized as more
questions are asked about the same slide. The average runtime decreases from
94.3\,s for a single query to 5.9\,s per query over the question sequence,
approaching an online reasoning cost of 2.73\,s per query. We restrict the
comparison to generative VLMs and WSI assistants with comparable
conversational inference interfaces; representation-oriented models such as
TITAN and agentic methods whose multi-round computation occurs internally
within a single query are excluded. As shown in Fig.~\ref{fig:efficiency}(b),
\SlideBank has higher consistency and lower runtime
than baselines.

%% file: tables/baseline.tex
\begin{table*}[t]
\centering
\caption{
\textbf{Standard WSI question-answering performance.}
We report standard single-query performance to examine whether this reusable representation preserves sufficient information for downstream WSI question answering. \textbf{Bold} and \underline{underline} show the best and second best result.}
\label{tab:baseline}
\resizebox{\textwidth}{!}{
\begin{tabular}{l c c c c c c}
\toprule
\textbf{Method} &
\textbf{Inference Input} &
\multicolumn{4}{c}{\textbf{SlideBench-BCNB Accuracy (\%) $\uparrow$}} &
\textbf{WSI-VQA (\%) $\uparrow$}
\\
\cmidrule(lr){3-6}
&
&
\textbf{Tumor Type} &
\textbf{Grading} &
\textbf{Subtype} &
\textbf{Weighted Avg.} &
\textbf{Accuracy}
\\
\midrule

\multicolumn{7}{l}{\textit{Vision-Language Models (Zero-shot)}} \\
Qwen2.5-VL-7B~\cite{bai2025qwen25vltechnicalreport}
    & Thumbnail & 58.13 & 27.11 & 16.07 & 34.06 & 33.43 \\
Qwen3-VL-8B~\cite{bai2025qwen3}
    & Thumbnail & 55.01 & 27.65 & 16.92 & 33.43 & 39.92 \\
Qwen3.5-4B~\cite{team2026qwen3}
    & Thumbnail & 39.89 & 38.34 & 17.30 & 31.56& 44.53\\
\midrule
LLaVA-Med~\cite{li2023llava}
    & Thumbnail & 26.56 & 27.75 & 22.40 & 25.48 & 31.87 \\
Quilt-LLaVA~\cite{seyfioglu2024quilt}
    & Thumbnail & 44.33 & 32.18 & 25.05 & 33.93 & 27.65 \\
PathGen-LLaVA~\cite{sun2024pathgen}
    & Thumbnail & 45.18 & 33.05 & 22.68 &	33.66 & 39.90 \\
Patho-R1~\cite{zhang2025pathor1}
    & Thumbnail &61.11 & 29.93& \underline{27.57} &39.95& 44.28 \\

\midrule
\multicolumn{7}{l}{\textit{WSI-trained Models}} \\
WSI-LLaVA-7B~\cite{wsillava2025iccv}
    & Slide & 42.53 & 30.02 & \textbf{32.42} & 35.21 & 49.61 \\
TITAN~\cite{ding2025multimodal}
    & Slide & 83.55 & 27.00 & 21.27 & 44.67 & 43.12 \\
SlideChat~\cite{chen2025slidechat}
    & Slide & 87.99 & 17.86 & 20.41 & 43.14 & \underline{52.05} \\

\midrule
\multicolumn{7}{l}{\textit{Agentic Systems}} \\
MedAgents~\cite{tang2024medagents}
    & Thumbnail & \underline{89.38} & 37.37 & 15.12 & 	47.72 & 42.75 \\
PathAgent~\cite{chen2025pathagent}
    & Slide & 50.76& \textbf{46.00}& 27.08& 41.08& \underline{52.05}\\

\midrule
\SlideBank (w/ Quilt-LLaVA)
    & Thumbnail + Evidence
    & \textbf{89.82}
    & 33.56
    & 27.22
    & \textbf{50.92}
    & 43.75 \\
\SlideBank (w/ Patho-R1)
    & Thumbnail + Evidence
    & 83.27
    & \underline{40.06}
    & 26.47
    & \underline{50.36}
    & \textbf{52.77}   \\

\bottomrule
\end{tabular}
}
\end{table*}

%% file: tables/ablation.tex
\begin{table}[t]
    \centering
    \caption{Ablation on hierarchical evidence sources and evidence access strategies (w/ Patho-R1).}
    \label{tab:ablation}

    \setlength{\tabcolsep}{3pt}

    \begin{tabular}{l c cc}
        \toprule
        \textbf{Variant}
        & \textbf{Evidence}
        & \textbf{WSI-VQA}
        & \textbf{BCNB} \\
        \midrule

        \multicolumn{4}{l}{\emph{Evidence hierarchy}} \\
        Global only         & G           & 48.55 & 40.57 \\
        Anchor only         & A           & 51.98 & 50.23 \\
        Patch only          & P           & 51.45 & 48.06 \\
        Full hierarchy      & G+A+P       & \textbf{52.77} & \textbf{50.36} \\

        \midrule

        \multicolumn{4}{l}{\emph{Evidence access}} \\
        Random        & G+A+P       & 50.13 & 49.62 \\
        Signal-guided & G+A+P       & \textbf{52.77} & \textbf{50.36} \\

        \bottomrule
    \end{tabular}
\end{table}

%% file: tables/consistency.tex
\begin{table}[t]
    \centering
    \caption{Performance comparison on WSI-VQA under the multi-turn setting. Rephrase Cons. measures the prediction
    consistency across semantically equivalent question formulations. }
    \label{tab:multiturn}

     \setlength{\tabcolsep}{3pt}

    \begin{tabular}{l ccc}
        \toprule
        \textbf{Method}
        & \textbf{Acc $\uparrow$}
        & \textbf{$\Delta$Acc }
        & \textbf{Rephrase Cons. $\uparrow$} \\
        \midrule

        LLaVA-Med \cite{li2023llava}
        & 16.18 & -15.69 & 62.83 \\

        Quilt-LLaVA \cite{seyfioglu2024quilt}
        & 25.22 & -2.43 & 84.25 \\

        PathGen-LLaVA \cite{sun2024pathgen}
        & 23.06 & -16.84 & 87.27 \\

        Patho-R1 \cite{zhang2025pathor1}
        & 41.93 & -2.35 & 80.94 \\

        \midrule

        WSI-LLaVA \cite{wsillava2025iccv}
        & 17.06 & -32.55 & 73.21 \\

        SlideChat \cite{chen2025slidechat}
        & 42.50 & -9.55 & 88.69 \\

        \midrule

        \SlideBank (w/ Quilt-LLaVA)
        & 42.37
        & -1.38
        & 99.47 \\

        \SlideBank (w/ Patho-R1)
        & 48.94
        & -3.83
        & 99.21 \\

        \bottomrule
    \end{tabular}%
    
\end{table}

%% file: sec/5_conclusion.tex
\section{Conclusion}
We introduced \SlideBank, a training-free framework that converts each WSI into a persistent, concept-indexed, and spatially grounded evidence bank. \SlideBank separates slide exploration from question answering, organizes multi-scale morphology through pathology signals, and retrieves question-relevant evidence with explicit links to supporting regions and patches. Experiments on WSI-VQA and SlideBench-BCNB show competitive performance, improved stability across repeated queries, and efficiency gains from evidence reuse. These results highlight structured evidence organization as a promising direction for WSI reasoning.

%% file: sec/appendix.tex
\appendix

\section{Prompts}
\label{app:prompts}

\subsection{Hierachical Evidence Bank Construction Prompts}
This section lists the complete prompts used by the offline bank-construction
agent (Sec.~\ref{sec:bank}). All prompts are issued to the same frozen
pathology VLM backbone that is used for online reasoning. Decoding is greedy with at most 128 new tokens. The VLM infers three times during the construction: anchor localization inside a retained tile (P1), view selection inside an anchor context (P2 and P3), and multi-scale view description (P4).

\paragraph{P1: anchor localization within a tile.}
The tile is resized to $256{\times}256$ and partitioned into a
$4{\times}4$ grid. The agent scores all 16 cells and returns the top
\texttt{\{target\_count\}} cells, which is 3 for tumor-like tiles and 2
otherwise. P1 shows the anchor localization prompt.

\begin{tcolorbox}[
    enhanced,
    breakable,
    title={P1: Anchor Localization Prompt},
    colback=gray!3,
    colframe=black!60,
    boxrule=0.6pt,
    arc=1mm,
    left=2mm,
    right=2mm,
    top=1.5mm,
    bottom=1.5mm,
    fonttitle=\bfseries,
]
\footnotesize
\begin{Verbatim}[breaklines=true, breaksymbolleft=]
You are a pathology diagnostic agent performing focused anchor planning.
Input image is one low-resolution tile (256x256) from a WSI.
Split it into a 4x4 grid (rows 0..3, cols 0..3).
Select TOP-{target_count} cells for anchor extraction.
Prioritize diagnostically rich structure: tumor-like atypia, hypercellularity,
  necrosis interface, and gland/duct architecture.
Evaluate all 16 cells before choosing.
Use ZERO-BASED indices only: row,col must be in [0,1,2,3].
Do NOT default to corner cells (especially R0C0) unless there is clear strongest
  evidence there.
If multiple cells are close in quality, prefer the one closer to the tile center
  over corner cells.
Return EXACTLY one JSON object and nothing else.
Output format requirements:
1) Key: selected_cells
2) selected_cells length MUST be exactly {target_count}
3) Each item must include: row (int 0..3), col (int 0..3), confidence (float 0..1),
   rationale (short text)
4) selected_cells must contain unique (row,col) only
Do NOT copy template text from this prompt.
Do NOT output placeholder values such as "...", "TBD", or dummy coordinates.
Coordinates must come from visual evidence in the image.
Rationale must mention at least one concrete visual cue (e.g., pleomorphism,
  nuclear crowding, necrosis, tubule/gland structure).
No tool_trace key. No markdown. No explanation text.
tile_id={tile_id}; coarse_label_hint={coarse_label}; tissue_ratio={tissue_ratio};
  target_count={target_count}.
Keep rationale short.
\end{Verbatim}
\end{tcolorbox}

\paragraph{P2: fine-detail view selection within an anchor.}
The anchor region is resized to
$256{\times}256$ and again partitioned into a $4{\times}4$ grid. The agent first
scores all 16 cells with a quality-control label and morphology tags, and then
commits to \texttt{\{target\_count\}} cells that are extracted as 512-pixel
level-0 crops. The tag vocabulary is the image-side entry point of the signal ontology. We show the  example of the breast used for WSI-VQA and BCNB in P2. The corresponding tag list
can be replaced by the ontology of the target cohort for other organs. 

\begin{tcolorbox}[
    enhanced,
    breakable,
    title={P2: Fine-Detail View Selection Prompt ($20\times$)},
    colback=gray!3,
    colframe=black!60,
    boxrule=0.6pt,
    arc=1mm,
    left=2mm,
    right=2mm,
    top=1.5mm,
    bottom=1.5mm,
    fonttitle=\bfseries,
]
\footnotesize
\begin{Verbatim}[breaklines=true, breaksymbolleft=]
You are a pathology diagnostic agent selecting evidence-driven patches.
Input: 4096x4096 level-0 context resized to 256x256. Split into 4x4 grid (row/col 0..3).
Select TOP-{target_count} cells for extraction.

Priority is morphology evidence, not geometry. Do NOT optimize for diagonals/symmetry.
Spatial diversity is ONLY a tie-breaker when evidence is similar.

Focus signals (summarized):
- High-grade invasive tumor cues: marked nuclear pleomorphism, hypercellular dark
  hotspots (mitoses proxy), solid growth / low tubules.
- DCIS cues: duct-centered intraductal proliferation; solid/cribriform patterns;
  comedo-type central necrosis.
- Necrosis and calcifications: pale granular debris/ghost areas; tiny bright
  punctate clusters.
- LVI / node mets: tumor clusters within vessel-like spaces or in nodal tissue
  (select only if clearly suggested).
- Margin involvement/close: tumor abutting/near tissue edge (only if tumor is
  visibly near boundary).

Return EXACTLY one JSON object.
Keys:
1) cell_scores (len 16): each {row,col,qc:"ok"|"blur"|"artifact",
score:0..1,tags:[...],note}
   - tags choose from: pleomorphism, mitosis_hotspot, low_tubules_solid, dcis,
     dcis_comedo, dcis_cribriform_solid, lvi_suspect, microcalc, necrosis,
     margin_close, margin_involved, node_met, micro_met
   - note: <=12 words, visible cue only (no diagnosis).
2) selected_cells (len {target_count}): each {row,col,conf:0..1,primary_tag,rationale}
   - rationale: <=12 words, include primary_tag + one visible cue.

Rules:
- unique (row,col). Avoid qc=artifact unless insufficient.
- do not select score<0.35 unless not enough cells >=0.35.
- rank by score. tie (diff<0.08): prefer qc ok, then cover different rows/cols.
- never pick solely for spatial spread.

tile_id={tile_id}; anchor_id={anchor_id}; coarse_label_hint={coarse_label};
  target_count={target_count}.
\end{Verbatim}
\end{tcolorbox}

\paragraph{P3: architectural context view selection.}
Each anchor additionally receives one independently selected low-magnification
view ($10\times$). Selection is deliberately decoupled from P2 so that
the context view is not forced to coincide with the cell-level evidence, and the
prompt asks for architectural cues such as transition fronts, necrosis
interfaces, and ductal layout.

\begin{tcolorbox}[
    enhanced,
    breakable,
    title={P3: Architectural Context View Selection Prompt ($10\times$)},
    colback=gray!3,
    colframe=black!60,
    boxrule=0.6pt,
    arc=1mm,
    left=2mm,
    right=2mm,
    top=1.5mm,
    bottom=1.5mm,
    fonttitle=\bfseries,
]
\scriptsize
\begin{Verbatim}[breaklines=true, breaksymbolleft=]
You are a pathology diagnostic agent selecting
low-magnification context patches.
Input: anchor context resized to 256x256 and split
into 4x4 grid (row/col 0..3).
Select TOP-{target_count} cells for ~10x context
extraction.

Goal:
- Capture broader tissue architecture and lesion
  context, independently from high-magnification
  picks.
- Prioritize representative context zones
  (transition fronts, heterogeneous structure,
  necrosis interface, ductal layout).
- Avoid selecting solely by corners/diagonals/
  symmetry.

Return EXACTLY one JSON object.
Keys:
1) selected_cells (len {target_count}): each
   {row,col,confidence,rationale}
2) rationale must mention low-mag context cue
   (architecture/interface/distribution)
   in <=12 words.

Rules:
- Use unique (row,col).
- Evaluate all 16 cells before selecting.
- Do NOT copy text from prompt.
- No markdown, no tool_trace, no extra keys.
tile_id={tile_id}; anchor_id={anchor_id};
  coarse_label_hint={coarse_label};
  target_count={target_count}.
\end{Verbatim}
\end{tcolorbox}

\paragraph{P4: multi-scale view description.}
Every extracted view is described independently, producing the triple
$(d_v,c_v,z_v)$ of Sec.~\ref{sec:bank}. The prompt constrains the model to observable
morphology and to a fixed JSON schema, and it states the specimen organ
\texttt{\{organ\}} of the cohort under evaluation (e.g., breast for WSI-VQA and
BCNB) to suppress the organ drift that pathology VLMs exhibit on isolated
fields. Diagnostic conclusions, grading terms, and management statements are
excluded at the prompt level and again removed by the post-processing rules
described below, so that the bank stores observations.

\begin{tcolorbox}[
    enhanced,
    breakable,
    title={P4: Multi-Scale View Description Prompt},
    colback=gray!3,
    colframe=black!60,
    boxrule=0.6pt,
    arc=1mm,
    left=2mm,
    right=2mm,
    top=1.5mm,
    bottom=1.5mm,
    fonttitle=\bfseries,
]
\scriptsize
\begin{Verbatim}[breaklines=true, breaksymbolleft=]
You are analyzing ONE {magnification} H&E patch from a human {organ} specimen.
Describe only visible morphology; do not infer another organ site, diagnosis, subtype, grade, biomarkers, or treatment. Return exactly one JSON object with this schema and no other keys or text:
{"caption":"2-3 morphology-only sentences",
 "confidence":0.75,"qc_flags":["none"]}

caption:
- 2-3 sentences of concise pathologist-style morphology description (<=100 words).
- Start with concrete visible features: color, cellularity, nuclear features, architecture, stroma, necrosis, calcifications, edge.
- Use only observable morphology (e.g., enlarged irregular nuclei, solid nests, duct-like lumen, central pale debris, bright puncta, tumor at edge).
- No diagnostic summary. No grading terms. No signal names. No recommendations.
- Any high-level inference must be explicitly tied to a stated visible feature.
- If minimal tissue or no clear morphology -> brief description + qc_flag='low_tissue'.

confidence:
- >=0.8 if multiple clear concrete features.
- 0.4-0.7 if limited but interpretable morphology.
- <=0.3 if sparse, blurred, or uncertain.

qc_flags: [none, blur, out_of_focus, artifact, low_tissue].

tile_id={tile_id}; anchor_id={anchor_id}; view_id={view_id}.
\end{Verbatim}
\end{tcolorbox}

\subsection{Inference Prompts}
\label{app:prompt_routing}

\paragraph{Size-question specialization.}
Tumor extent is the one question family that is answered from measurements
stored in the global record. A short prompt (P5) decides whether the current question is a measurement question, and if it is, the question is answered from the thumbnail with the scale prior of the stored pyramid geometry (P6). Questions that are not measurement questions are ignored by this branch and follow the ordinary routing path.

\begin{tcolorbox}[
    enhanced,
    breakable,
    title={P5: Size-Question Router Prompt},
    colback=gray!3,
    colframe=black!60,
    boxrule=0.6pt,
    arc=1mm,
    left=2mm,
    right=2mm,
    top=1.5mm,
    bottom=1.5mm,
    fonttitle=\bfseries,
]
\scriptsize
\begin{Verbatim}[breaklines=true, breaksymbolleft=]
You are a question router for pathology MCQ.
Determine whether this is primarily a tumor
size/dimension measurement question.
Size-related means it asks about size, dimension,
diameter, extent, or values in mm/cm.
Return strict JSON only:
{
  "is_size_question": true,
  "reason": "short reason"
}

Question:
{question}

Options:
{options}
\end{Verbatim}
\end{tcolorbox}

\begin{tcolorbox}[
enhanced,
    breakable,
    title={P6: Size Answering Prompt},
    colback=gray!3,
    colframe=black!60,
    boxrule=0.6pt,
    arc=1mm,
    left=2mm,
    right=2mm,
    top=1.5mm,
    bottom=1.5mm,
    fonttitle=\bfseries,
]
\scriptsize
\begin{Verbatim}[breaklines=true, breaksymbolleft=]
You are a pathology multiple-choice assistant.
This question is size-related. Use the thumbnail
image to choose one option.
Scale prior:
- Original WSI long side is about 50000 px.
- Thumbnail long side is about 1000 px
  (about 50x compression).
You may estimate lesion span on thumbnail then
multiply by about 50 when needed.
Return strict JSON only:
{
  "answer": "A",
  "reasoning": "short reason"
}

Question:
{question}

Options:
{options}
\end{Verbatim}
\end{tcolorbox}

\paragraph{Branch answering.}
Every retrieved item is evaluated independently by the same VLM with the
level-specific prompt of P7 (a-c). The three variants differ
only in which memory content accompanies the image: the global branch sees the
slide thumbnail alone, the anchor branch additionally receives the stored anchor
description, and the patch branch receives the stored view caption. The final answer is from the level-weighted consensus.
\tcbset{
    branchprompt/.style={
        enhanced,
        breakable,
        colback=gray!3,
        colframe=black!60,
        boxrule=0.6pt,
        arc=1mm,
        left=2mm,
        right=2mm,
        top=1.5mm,
        bottom=1.5mm,
        fonttitle=\bfseries,
    }
}

\begin{tcolorbox}[
    branchprompt,
    title={P7a: Global Branch Answering Prompt}
]
\footnotesize
\begin{Verbatim}[breaklines=true, breaksymbolleft=]
You are a pathology multiple-choice assistant.
Given one pathology slide thumbnail image and one multiple-choice question,
choose exactly one option.
Respond with only one letter: A, B, C, or D.

Question:
{question}

Options:
{options}

Answer:
\end{Verbatim}
\end{tcolorbox}

\begin{tcolorbox}[
    branchprompt,
    title={P7b: Anchor Branch Answering Prompt}
]
\footnotesize
\begin{Verbatim}[breaklines=true, breaksymbolleft=]
You are a pathology multiple-choice assistant.
Given one pathology anchor-region image, its stored anchor description, and one
multiple-choice question, choose exactly one option.
Respond with only one letter: A, B, C, or D.

Anchor description:
{anchor_description}

Question:
{question}

Options:
{options}

Answer:
\end{Verbatim}
\end{tcolorbox}

\begin{tcolorbox}[
    branchprompt,
    title={P7c: Patch Branch Answering Prompt}
]
\footnotesize
\begin{Verbatim}[breaklines=true, breaksymbolleft=]
You are a pathology multiple-choice assistant.
Given one pathology patch image, its stored patch description, and one
multiple-choice question, choose exactly one option.
Respond with only one letter: A, B, C, or D.

Patch description:
{patch_description}

Question:
{question}

Options:
{options}

Answer:
\end{Verbatim}
\end{tcolorbox}

\paragraph{Multi-turn context.}
For multi-turn evaluation, the bounded history $\mathcal{H}_t$ is rendered as a
plain block that is prepended to the current question before the branch prompts
are built, so that each branch is conditioned on the same discourse context
while the visual evidence is retrieved afresh from the persistent bank. The
history contains predicted answers, not ground truth, and is truncated to the
most recent turns for the same slide.

\begin{tcolorbox}[
    title={P8: Multi-Turn Context Block},
    colback=gray!3,
    colframe=black!60,
    boxrule=0.6pt,
    arc=1mm,
    left=2mm,
    right=2mm,
    top=1.5mm,
    bottom=1.5mm,
    fonttitle=\bfseries,
]
\scriptsize
\begin{verbatim}
Previous QA context for this slide:
Q1: {question_1}
A1: {predicted_answer_1}
...
Qh: {question_h}
Ah: {predicted_answer_h}

Current question:
{question}
\end{verbatim}
\end{tcolorbox}

\section{Pathology Concept and Signal Ontology}
\label{app:ontology}

SlideBank uses a compact pathology ontology to connect a clinical question with
relevant visual evidence. A \emph{concept} describes the clinical intent of a
question (e.g., tumor grade or lymphovascular invasion), whereas a \emph{signal}
describes an atomic morphologic finding that can be assessed in a local image
region (e.g., high mitotic activity). The resulting reasoning pipeline is

\[
q \longrightarrow c(q) \longrightarrow \mathcal{S}_{c(q)}
\longrightarrow \mathcal{E}_{q} \longrightarrow \hat{a},
\]

where a question $q$ is first assigned to a concept $c(q)$. The concept selects
a set of relevant signals $\mathcal{S}_{c(q)}$, which are then used to retrieve
multi-scale evidence $\mathcal{E}_{q}$ for answer generation. This separation
makes the retrieval process interpretable: each answer can be traced from the
question concept to the supporting morphologic signals and image regions.

\subsection{Concept Vocabulary}
\label{app:concept_vocabulary}

The concept vocabulary covers common morphology-centered questions in breast
pathology, together with metadata-oriented and morphology-based proxy tasks.
Table~\ref{tab:concept_signal_mapping} summarizes the concepts and their
associated signals. The mapping is many-to-many because a single clinical
concept may depend on several morphologic findings, and the same finding may
support more than one concept. Concepts without a dedicated signal are answered
using global and multi-scale regional evidence.

\begin{table*}[t]
\centering
\caption{Clinical concepts and their associated morphologic signals.}
\label{tab:concept_signal_mapping}
\scriptsize
\setlength{\tabcolsep}{4pt}
\begin{tabularx}{\textwidth}{@{}l p{0.35\textwidth} X@{}}
\toprule
ID & Concept & Associated signals \\
\midrule
M01 & Tumor presence and distribution & S104 \\
M02 & Histological tumor type (coarse) & S103, S111, S113 \\
M03 & Tumor grade & S101, S102, S103, S104 \\
M04 & Necrosis assessment & S112, S132 \\
M05 & Inflammation / tumor-infiltrating lymphocytes & No dedicated signal; multi-scale evidence \\
M06 & In-situ component (DCIS-like) & S111, S112, S113 \\
M07 & Margin status & S141, S142 \\
M08 & Lymphovascular invasion & S121 \\
M09 & Perineural invasion & No dedicated signal; multi-scale evidence \\
M10 & Microcalcifications & S131 \\
M11 & Benign or non-neoplastic background changes & No dedicated signal; multi-scale evidence \\
MD01 & Specimen, procedure, and site metadata & Global evidence \\
MD02 & Tumor size, extent, and stage elements & S141 (weak proxy), with global evidence \\
P01 & Biomarker or outcome proxy prediction & S101--S104, S111, S113, S132 \\
\bottomrule
\end{tabularx}
\end{table*}

\subsection{Concept-to-signal Mapping}
\label{app:signal_vocabulary}

Signals are affirmative, locally assessable morphologic findings rather than
diagnostic labels. Table~\ref{tab:signal_vocabulary} lists the signal vocabulary
used by SlideBank. Each signal is linked to the image regions in which it is
observed, allowing the same evidence to be reused across related concepts.

\begin{table*}[t]
\centering
\caption{Atomic morphologic signal vocabulary.}
\label{tab:signal_vocabulary}
\scriptsize
\setlength{\tabcolsep}{4pt}
\begin{tabularx}{\textwidth}{@{}l p{0.31\textwidth} X@{}}
\toprule
ID & Signal & Morphologic cue \\
\midrule
S101 & High nuclear pleomorphism & Marked variation in nuclear size and shape. \\
S102 & High mitotic activity & Frequent mitotic figures indicating increased proliferation. \\
S103 & Low tubule formation & Reduced tubular differentiation with predominantly solid growth. \\
S104 & Support for high tumor grade & Combined visual evidence consistent with a higher-grade appearance. \\
S111 & DCIS present & In-situ ductal proliferation without clear stromal invasion. \\
S112 & DCIS with comedo necrosis & Intraluminal necrotic debris compatible with comedo morphology. \\
S113 & Solid or cribriform DCIS architecture & Solid or cribriform in-situ architecture within ductal units. \\
S121 & Lymphovascular invasion present & Suspicious tumor emboli within endothelial-lined spaces. \\
S131 & Microcalcification present & Fine basophilic calcific deposits in tumor-associated tissue. \\
S132 & Tumor necrosis present & Tumor-associated necrosis with loss of viable architecture. \\
S141 & Tumor close to margin & Tumor located close to a specimen edge in the sampled context. \\
S142 & Tumor involving margin & Tumor involving an edge or margin-like boundary. \\
S151 & Lymph-node metastasis present & Metastatic epithelial cells in a lymphoid background. \\
S152 & Micrometastasis present & A small metastatic focus compatible with micrometastatic burden. \\
\bottomrule
\end{tabularx}
\end{table*}

For each evidence region, a signal is represented as \texttt{present},
\texttt{unknown}, or \texttt{absent}, together with a confidence score and its
supporting views. During retrieval, evidence linked to the target signals is
prioritized, while uncertain or negative observations may also be retained when
they are relevant to the question. 

\subsection{Multi-scale Evidence Organization}
\label{app:evidence_organization}

SlideBank organizes evidence at three complementary scales: a global view
captures slide-level context, anchor regions localize diagnostically informative
areas, and patch views preserve detailed architectural or cellular morphology.
Each signal is linked to its supporting anchor and patch views. Consequently,
the evidence supplied to the vision-language model retains both its spatial
origin and its morphologic interpretation, enabling an answer to be traced back
to the corresponding signal and image region.
\section{Datasets and Baselines}
\label{app:evaluation}

\subsection{Dataset Statistics}
\label{app:dataset_statistics}

\paragraph{WSI-VQA.}
The standard evaluation set contains 388 four-option questions associated with
85 WSIs. Questions cover slide-level diagnosis and morphology as well as
attributes such as grade, size, receptor status, margins, and staging. We use
the multiple-choice conversion of the released annotations and preserve the
original slide--question association and option text. Accuracy is computed over
questions, so slides with more annotated questions contribute proportionally
more examples.

\paragraph{SlideBench-BCNB.}
The BCNB split contains 1,058 breast-cancer WSIs. Following the standard
SlideBench protocol, we evaluate the three tasks that are answerable from the
released multiple-choice annotations: tumor type (1,058 questions),
histological grading (926 questions), and molecular subtype (1,058 questions),
for 3,042 questions in total. Tumor-type and grading questions have three
answer candidates, while molecular-subtype questions have four. We report each
task separately and compute the overall BCNB result as the
question-count-weighted average
\begin{equation}
\begin{aligned}
  \operatorname{Acc}_{\mathrm{BCNB}}
  =\frac{1}{3042}\big(&1058\operatorname{Acc}_{\mathrm{tumor}}
  +926\operatorname{Acc}_{\mathrm{grade}}
  +1058\operatorname{Acc}_{\mathrm{subtype}}\big).
\end{aligned}
  \label{eq:bcnb_weighted_accuracy}
\end{equation}

\subsection{Baselines}
\label{app:baselines}

All methods are evaluated on the same questions and answer candidates. Unless
otherwise specified, we follow the released inference protocol for each
baseline.

\noindent\textbf{General-purpose and domain-specif{}ic VLMs.} 
We evaluate Qwen2.5-VL-7B \cite{bai2025qwen25vltechnicalreport}, Qwen3-VL-8B \cite{bai2025qwen3}, and Qwen3.5-4B \cite{team2026qwen3} as general-purpose
VLMs, together with LLaVA-Med \cite{li2023llava}, Quilt-LLaVA \cite{seyfioglu2024quilt}, PathGen-LLaVA \cite{sun2024pathgen}, and Patho-R1 \cite{zhang2025pathor1} as
medical- or pathology-tuned VLMs. Each model receives a single
$1024\!\times\!1024$ WSI thumbnail and the multiple-choice question, without
access to additional high-resolution fields.

\noindent\textbf{WSI-LLaVA \cite{wsillava2025iccv}.}
WSI-LLaVA is a multimodal model designed for gigapixel WSI understanding. It
uses hierarchical slide representations and a multi-stage alignment and
instruction-tuning strategy to connect WSI morphology with language.

\noindent\textbf{TITAN \cite{ding2025multimodal}.}
TITAN is a multimodal whole-slide foundation model pretrained through visual
self-supervision and vision--language alignment. It encodes an entire WSI into
a slide-level representation that can be transferred to downstream pathology
tasks without task-specific fine-tuning.

\noindent\textbf{SlideChat \cite{chen2025slidechat}.}
SlideChat is a vision--language assistant developed specifically for
gigapixel whole-slide pathology. It is instruction-tuned on slide-level
captions and VQA pairs to support WSI description and question answering.

\noindent\textbf{MedAgents \cite{tang2024medagents}.}
MedAgents is a multi-agent medical reasoning framework in which specialized
agents analyze a question from complementary perspectives and aggregate their
conclusions. We include it to assess whether inference-time collaboration can
improve WSI question answering without a dedicated evidence-retrieval module.We adapt the MedAgents multi-agent collaboration framework to the multimodal
setting using GPT-4o as the backbone. Each agent receives the WSI thumbnail
and the multiple-choice question, and the final answer is obtained through
multi-round specialist discussion and consensus.

\noindent\textbf{PathAgent \cite{chen2025pathagent}.}
PathAgent is a training-free agentic framework for WSI analysis. It iteratively
navigates the slide, extracts morphological evidence at selected regions and
magnifications, and integrates the observations to produce an interpretable
answer.For the backbone setting, we use the official PathAgent implementation.

\section{Additional Implementation Details}
\label{app:implementation_details}

\noindent\textbf{Fusion weights.}
We use fixed fusion weights
$(\lambda_G,\lambda_A,\lambda_P)=(0.2,0.4,0.4)$ across all datasets and
backbones, without dataset-specific tuning.

\noindent\textbf{Multi-turn evaluation.}
For each WSI-VQA question, Gemini-2.5-Flash generates five semantically
equivalent rephrasings, and the answer options are independently shuffled while
preserving the correct option text. After retaining complete six-form groups,
the evaluation set contains 382 source questions and 2,292 turns from 80 WSIs,
with an average of 29 turns per WSI. The evidence bank is constructed once per
slide and remains fixed throughout the conversation; each turn uses the
immediately preceding
predicted question--answer pair as context, without access to ground-truth
answers. Rephrase consistency is computed after mapping the shuffled option
letters back to their semantic option text.

\noindent\textbf{Answer parsing.}
We use a deterministic parser that first extracts a valid option letter from a
JSON answer field or an explicit answer expression. If this fails, it accepts
a standalone option letter or matches the normalized generated text to the
provided option text. Outputs that cannot be resolved to a valid candidate are
counted as incorrect, without using an additional language model for
adjudication.

%% file: reference.bib
@String(CVPR= {IEEE Conf. Comput. Vis. Pattern Recog.})

@String(ICCV= {Int. Conf. Comput. Vis.})

@String(AAAI = {AAAI})

@String(CVPR  = {CVPR})

@String(ICCV  = {ICCV})

@inproceedings{wsillava2025iccv,
  title     = {WSI-LLaVA: A Multimodal Large Language Model for Whole Slide Image},
  author    = {Liang, Yuci and Lyu, Xinheng and Chen, Wenting and Ding, Meidan and Zhang, Jipeng and He, Xiangjian and Wu, Song and Xing, Xiaohan and Yang, Sen and Wang, Xiyue and Shen, Linlin},
  booktitle = {Proceedings of the IEEE/CVF International Conference on Computer Vision (ICCV)},
  month     = {October},
  year      = {2025},
  pages     = {22718--22727}
}

@article{giant2025arxiv,
  title={Navigating gigapixel pathology images with large multimodal models},
  author={Buckley, Thomas A and Weihrauch, Kian R and Latham, Katherine and Zhou, Andrew Z and Manrai, Padmini A and Manrai, Arjun K},
  journal={arXiv preprint arXiv:2511.19652},
  year={2025}
}

@inproceedings{pathfinder2025arxiv,
  title     = {PathFinder: A Multi-Modal Multi-Agent System for Medical Diagnostic Decision-Making Applied to Histopathology},
  author    = {Ghezloo, Fatemeh and Seyfioglu, Mehmet Saygin and Soraki, Rustin and Ikezogwo, Wisdom O. and Li, Beibin and Vivekanandan, Tejoram and Elmore, Joann G. and Krishna, Ranjay and Shapiro, Linda},
  booktitle = {Proceedings of the IEEE/CVF International Conference on Computer Vision (ICCV)},
  month     = {October},
  year      = {2025},
  pages     = {23431--23441}
}

@article{slideseek2025arxiv,
  title={Evidence-based diagnostic reasoning with multi-agent copilot for human pathology},
  author={Weishaupt, Luca L and Chen, Chengkuan and Williamson, Drew FK and Chen, Richard J and Jaume, Guillaume and Ding, Tong and Chen, Bowen and Vaidya, Anurag and Le, Long Phi and Lu, Ming Y and others},
  journal={arXiv preprint arXiv:2506.20964},
  year={2025}
}

@inproceedings{chen2025slidechat,
  title={Slidechat: A large vision-language assistant for whole-slide pathology image understanding},
  author={Chen, Ying and Wang, Guoan and Ji, Yuanfeng and Li, Yanjun and Ye, Jin and Li, Tianbin and Hu, Ming and Yu, Rongshan and Qiao, Yu and He, Junjun},
  booktitle={Proceedings of the Computer Vision and Pattern Recognition Conference},
  pages={5134--5143},
  year={2025}
}

@inproceedings{seyfioglu2024quilt,
  title={Quilt-llava: Visual instruction tuning by extracting localized narratives from open-source histopathology videos},
  author={Seyfioglu, Mehmet Saygin and Ikezogwo, Wisdom O and Ghezloo, Fatemeh and Krishna, Ranjay and Shapiro, Linda},
  booktitle={Proceedings of the IEEE/CVF Conference on Computer Vision and Pattern Recognition},
  pages={13183--13192},
  year={2024}
}

@inproceedings{chen2024wsi,
  title={Wsi-vqa: Interpreting whole slide images by generative visual question answering},
  author={Chen, Pingyi and Zhu, Chenglu and Zheng, Sunyi and Li, Honglin and Yang, Lin},
  booktitle={European Conference on Computer Vision},
  pages={401--417},
  year={2024},
  organization={Springer}
}

@article{bai2025qwen3,
  title={Qwen3-vl technical report},
  author={Bai, Shuai and Cai, Yuxuan and Chen, Ruizhe and Chen, Keqin and Chen, Xionghui and Cheng, Zesen and Deng, Lianghao and Ding, Wei and Gao, Chang and Ge, Chunjiang and others},
  journal={arXiv preprint arXiv:2511.21631},
  year={2025}
}

@inproceedings{sun2024pathgen,
  title={Pathgen-1.6 m: 1.6 million pathology image-text pairs generation through multi-agent collaboration},
  author={Sun, Yuxuan and Zhang, Yunlong and Si, Yixuan and Zhu, Chenglu and Zhang, Kai and Shui, Zhongyi and Li, Jingxiong and Gong, Xuan and Lyu, Xinheng and Lin, Tao and others},
  booktitle={International Conference on Learning Representations},
  volume={2025},
  pages={94611--94653},
  year={2025}
}

@misc{bai2025qwen25vltechnicalreport,
      title={Qwen2.5-VL Technical Report}, 
      author={Shuai Bai and Keqin Chen and Xuejing Liu and Jialin Wang and Wenbin Ge and Sibo Song and Kai Dang and Peng Wang and Shijie Wang and Jun Tang and Humen Zhong and Yuanzhi Zhu and Mingkun Yang and Zhaohai Li and Jianqiang Wan and Pengfei Wang and Wei Ding and Zheren Fu and Yiheng Xu and Jiabo Ye and Xi Zhang and Tianbao Xie and Zesen Cheng and Hang Zhang and Zhibo Yang and Haiyang Xu and Junyang Lin},
      year={2025},
      eprint={2502.13923},
      archivePrefix={arXiv},
      primaryClass={cs.CV},
      url={https://arxiv.org/abs/2502.13923}, 
}

@article{lyu2025wsi,
  title={Wsi-agents: A collaborative multi-agent system for multi-modal whole slide image analysis},
  author={Lyu, Xinheng and Liang, Yuci and Chen, Wenting and Ding, Meidan and Yang, Jiaqi and Huang, Guolin and Zhang, Daokun and He, Xiangjian and Shen, Linlin},
  journal={arXiv preprint arXiv:2507.14680},
  year={2025}
}

@article{ding2025multimodal,
  title={A multimodal whole-slide foundation model for pathology},
  author={Ding, Tong and Wagner, Sophia J and Song, Andrew H and Chen, Richard J and Lu, Ming Y and Zhang, Andrew and Vaidya, Anurag J and Jaume, Guillaume and Shaban, Muhammad and Kim, Ahrong and others},
  journal={Nature medicine},
  pages={1--13},
  year={2025},
  publisher={Nature Publishing Group US New York}
}

@inproceedings{tang2024medagents,
  title={Medagents: Large language models as collaborators for zero-shot medical reasoning},
  author={Tang, Xiangru and Zou, Anni and Zhang, Zhuosheng and Li, Ziming and Zhao, Yilun and Zhang, Xingyao and Cohan, Arman and Gerstein, Mark},
  booktitle={Findings of the Association for Computational Linguistics: ACL 2024},
  pages={599--621},
  year={2024}
}

@article{huang2025survagent,
  title={SurvAgent: Hierarchical CoT-Enhanced Case Banking and Dichotomy-Based Multi-Agent System for Multimodal Survival Prediction},
  author={Huang, Guolin and Chen, Wenting and Yang, Jiaqi and Lyu, Xinheng and Luo, Xiaoling and Yang, Sen and Xing, Xiaohan and Shen, Linlin},
  journal={arXiv preprint arXiv:2511.16635},
  year={2025}
}

@article{chen2025pathagent,
  title={Pathagent: Toward interpretable analysis of whole-slide pathology images via large language model-based agentic reasoning},
  author={Chen, Jingyun and Cai, Linghan and Wang, Zhikang and Huang, Yi and Jiang, Songhan and Huang, Shenjin and Wang, Hongpeng and Zhang, Yongbing},
  journal={arXiv preprint arXiv:2511.17052},
  year={2025}
}

@article{lu2024visual,
  title={A visual-language foundation model for computational pathology},
  author={Lu, Ming Y and Chen, Bowen and Williamson, Drew FK and Chen, Richard J and Liang, Ivy and Ding, Tong and Jaume, Guillaume and Odintsov, Igor and Le, Long Phi and Gerber, Georg and others},
  journal={Nature medicine},
  volume={30},
  number={3},
  pages={863--874},
  year={2024},
  publisher={Nature Publishing Group US New York}
}

@article{comanici2025gemini,
  title={Gemini 2.5: Pushing the frontier with advanced reasoning, multimodality, long context, and next generation agentic capabilities},
  author={Comanici, Gheorghe and Bieber, Eric and Schaekermann, Mike and Pasupat, Ice and Sachdeva, Noveen and Dhillon, Inderjit and Blistein, Marcel and Ram, Ori and Zhang, Dan and Rosen, Evan and others},
  journal={arXiv preprint arXiv:2507.06261},
  year={2025}
}

@article{li2023llava,
  title={Llava-med: Training a large language-and-vision assistant for biomedicine in one day},
  author={Li, Chunyuan and Wong, Cliff and Zhang, Sheng and Usuyama, Naoto and Liu, Haotian and Yang, Jianwei and Naumann, Tristan and Poon, Hoifung and Gao, Jianfeng},
  journal={Advances in Neural Information Processing Systems},
  volume={36},
  pages={28541--28564},
  year={2023}
}

@article{packer2023memgpt,
  title={Memgpt: Towards llms as operating systems},
  author={Packer, Charles and Wooders, Sarah and Lin, Kevin and Fang, Vivian and Patil, Shishir G and Stoica, Ion and Gonzalez, Joseph E},
  journal={arXiv preprint arXiv:2310.08560},
  year={2023}
}

@inproceedings{park2023generative,
  title={Generative agents: Interactive simulacra of human behavior},
  author={Park, Joon Sung and O'Brien, Joseph and Cai, Carrie Jun and Morris, Meredith Ringel and Liang, Percy and Bernstein, Michael S},
  booktitle={Proceedings of the 36th annual acm symposium on user interface software and technology},
  pages={1--22},
  year={2023}
}

@inproceedings{chen2022scaling,
  title={Scaling vision transformers to gigapixel images via hierarchical self-supervised learning},
  author={Chen, Richard J and Chen, Chengkuan and Li, Yicong and Chen, Tiffany Y and Trister, Andrew D and Krishnan, Rahul G and Mahmood, Faisal},
  booktitle={Proceedings of the IEEE/CVF conference on computer vision and pattern recognition},
  pages={16144--16155},
  year={2022}
}

@inproceedings{hou2022h,
  title={H\^{} 2-MIL: exploring hierarchical representation with heterogeneous multiple instance learning for whole slide image analysis},
  author={Hou, Wentai and Yu, Lequan and Lin, Chengxuan and Huang, Helong and Yu, Rongshan and Qin, Jing and Wang, Liansheng},
  booktitle={Proceedings of the AAAI conference on artificial intelligence},
  volume={36},
  number={1},
  pages={933--941},
  year={2022}
}

@article{xu2025mem,
  title={A-mem: Agentic memory for llm agents},
  author={Xu, Wujiang and Liang, Zujie and Mei, Kai and Gao, Hang and Tan, Juntao and Zhang, Yongfeng},
  journal={arXiv preprint arXiv:2502.12110},
  year={2025}
}

@article{zhang2025g,
  title={G-memory: Tracing hierarchical memory for multi-agent systems},
  author={Zhang, Guibin and Fu, Muxin and Wan, Guancheng and Yu, Miao and Wang, Kun and Yan, Shuicheng},
  journal={arXiv preprint arXiv:2506.07398},
  year={2025}
}

@inproceedings{koh2020conceptbottleneck,
  title     = {Concept Bottleneck Models},
  author    = {Koh, Pang Wei and Nguyen, Thao and Tang, Yew Siang and Mussmann, Stephen and Pierson, Emma and Kim, Been and Liang, Percy},
  booktitle = {Proceedings of the 37th International Conference on Machine Learning},
  series    = {Proceedings of Machine Learning Research},
  volume    = {119},
  pages     = {5338--5348},
  year      = {2020},
  publisher = {PMLR}
}

@inproceedings{yuksekgonul2023posthoc,
  title     = {Post-hoc Concept Bottleneck Models},
  author    = {Yuksekgonul, Mert and Wang, Maggie and Zou, James},
  booktitle = {International Conference on Learning Representations},
  year      = {2023}
}

@inproceedings{zhang2025pathor1,
  title={Patho-r1: A multimodal reinforcement learning-based pathology expert reasoner},
  author={Zhang, Wenchuan and Zhang, Penghao and Guo, Jingru and Cheng, Tao and Chen, Jie and Zhang, Shuwan and Zhang, Zhang and Yi, Yuhao and Bu, Hong},
  booktitle={Proceedings of the AAAI Conference on Artificial Intelligence},
  volume={40},
  number={33},
  pages={28418--28426},
  year={2026}
}

@inproceedings{histoselect2026cvpr,
  title     = {Act Like a Pathologist: Tissue-Aware Whole Slide Image Reasoning},
  author    = {Huang, Wentao and Lyu, Weimin and Lou, Peiliang and Hu, Qingqiao and Hu, Xiaoling and Abousamra, Shahira and Han, Wenchao and Guo, Ruifeng and Zhou, Jiawei and Chen, Chao and Wang, Chen},
  booktitle = {Proceedings of the IEEE/CVF Conference on Computer Vision and Pattern Recognition (CVPR)},
  pages     = {6972--6981},
  year      = {2026},
  month     = {jun}
}

@article{alpaca2026naturecomm,
  title   = {{ALPaCA}: Adapting Llama for Pathology Context Analysis to Enable Slide-Level Question Answering},
  author  = {Gao, Zeyu and He, Kai and Su, Weiheng and Pang, Xiaobo and Machado, Ines P. and Jimenez-Linan, Mercedes and Rous, Brian and Wang, Chunbao and Li, Chengzu and McGough, William and Gao, Shangqi and Zhang, Di and Gong, Tieliang and Lu, Ming Y. and Mahmood, Faisal and Feng, Mengling and Li, Chen and Crispin-Ortuzar, Mireia},
  journal = {Nature Communications},
  year    = {2026},
  month   = {aug},
  doi     = {10.1038/s41467-026-76372-z}
}

@article{prism22026naturemed,
  title   = {End-to-End Multimodal Pathology Foundation Model with Clinical Dialogue},
  author  = {Vorontsov, Eugene and Shaikovski, George and Casson, Adam and Viret, Julian and Zimmermann, Eric and Tenenholtz, Neil and Wang, Yi Kan and Bernhard, Jan H. and Godrich, Ran A. and Retamero, Juan A. and Shia, Jinru and Gonen, Mithat and Weiser, Martin R. and Klimstra, David S. and Yousfi, Razik and Fusi, Nicol{\`o} and Fuchs, Thomas J. and Severson, Kristen and Liu, Siqi},
  journal = {Nature Medicine},
  year    = {2026},
  month   = {jul},
  doi     = {10.1038/s41591-026-04521-4}
}

@article{evipathbench2026arxiv,
  title   = {EviPathBench: Benchmarking Evidence Acquisition and Reasoning in Vision-Language Models for Whole-Slide Pathology},
  author  = {Liao, Dankai and Zhang, Tianyi and Wu, Yufeng and Zhang, Xinyue and Xue, Qiaochu and Liu, Zeyu and Zhao, Dachun and Cai, Linghan and Jin, Yueming},
  journal = {arXiv preprint arXiv:2607.19261},
  year    = {2026},
  month   = {jul},
  doi     = {10.48550/arXiv.2607.19261}
}

@article{beacon2026arxiv,
  title   = {Beyond Relevance: Bayesian Evidence Acquisition for Agentic Whole-Slide Image Reasoning},
  author  = {Wong, Bryan and Xu, Xun and Fu, Huazhu and Chen, Nancy F. and Yi, Mun Yong},
  journal = {arXiv preprint arXiv:2608.05757},
  year    = {2026},
  month   = {aug},
  doi     = {10.48550/arXiv.2608.05757}
}

@article{adaptivepath2026arxiv,
  title   = {Agentic Visual Reasoning in Whole-Slide Pathology Images via Active Perception},
  author  = {Chen, Jingyun and Liu, Fengchun and Cai, Linghan and Jiang, Songhan and Huang, Shenjin and Wang, Hongpeng and Yu, Lequan and Zhang, Yongbing},
  journal = {arXiv preprint arXiv:2608.08648},
  year    = {2026},
  month   = {aug},
  doi     = {10.48550/arXiv.2608.08648}
}

@inproceedings{alawode2026mllm,
  title={MLLM-HWSI: A Multimodal Large Language Model for Hierarchical Whole Slide Image Understanding},
  author={Alawode, Basit and Mahmood, Arif and Al Radi, Muaz Khalifa and Albastaki, Shahad and Khan, Asim and Bilal, Muhammad and Abdalla, Moshira Ali and Bennamoun, Mohammed and Javed, Sajid},
  booktitle={Proceedings of the IEEE/CVF Conference on Computer Vision and Pattern Recognition},
  pages={13732--13743},
  year={2026}
}

@article{yang2026pathnavigate,
  title={PathNavigate: A Training-Free Pathology Agent with Surprise-Guided Scan and Shared Slide Memory for Whole-Slide Image VQA},
  author={Yang, Chunze and Liu, Qidong and Zhao, Wenjie and Tang, Yue and Ge, Jiusong and Zhang, Di and Liu, Jiashuai and Wu, Lei and Lu, Junbo and Zhang, Ni and others},
  journal={arXiv preprint arXiv:2605.23559},
  year={2026}
}

@article{li2026pathmem,
  title={PathMem: Toward Cognition-Aligned Memory Transformation for Pathology MLLMs},
  author={Li, Jinyue and Liang, Yuci and Li, Qiankun and Lyu, Xinheng and Qian, Jiayu and Chen, Huabao and Wang, Kun and Zeng, Zhigang and Bharath, Anil Anthony and Liu, Yang},
  journal={arXiv preprint arXiv:2603.09943},
  year={2026}
}

@article{lu2024pathchat,
  title={A multimodal generative AI copilot for human pathology},
  author={Lu, Ming Y. and Chen, Bowen and Williamson, Drew F. K. and Chen, Richard J. and Zhao, Melissa and others},
  journal={Nature},
  volume={634},
  pages={466--473},
  year={2024},
  doi={10.1038/s41586-024-07618-3}
}

@misc{team2026qwen3,
    title  = {{Qwen3.5}: Towards Native Multimodal Agents},
    author = {{Qwen Team}},
    month  = {February},
    year   = {2026},
    url    = {https://qwen.ai/blog?id=qwen3.5}
}
